\pdfoutput=1

\documentclass[11pt]{article}

\usepackage{ACL2023}

\usepackage{times}
\usepackage{latexsym}

\usepackage[T1]{fontenc}

\usepackage[utf8]{inputenc}

\usepackage{microtype}

\usepackage{inconsolata}

\usepackage{amsmath, amssymb, amsfonts}
\usepackage{bm}
\usepackage{soul}
\usepackage{graphicx}
\usepackage{bbm}
\usepackage{multirow}
\usepackage{multicol}
\usepackage{booktabs}
\usepackage{enumitem}
\usepackage[inkscapelatex=false]{svg}
\usepackage{color}
\usepackage{soul}
\usepackage{pdfpages}
\usepackage{wrapfig}
\usepackage{url}
\newcommand{\sep}{\text{$\langle s \rangle$}}
\newcommand{\nline}{\text{$\langle n \rangle$}}

\newcommand{\nullrow}{\text{$\langle \varnothing \rangle$}}
\title{A Sequence-to-Sequence\&Set Model for Text-to-Table Generation}

\author{Tong Li\textsuperscript{1,2\footnotemark[1]}~~~
Zhihao Wang\textsuperscript{1,2\footnotemark[1]}~~~
Liangying Shao\textsuperscript{1}~~~ 
Xuling Zheng\textsuperscript{1,2\footnotemark[2]}~~~\\
\textbf{
Xiaoli Wang\textsuperscript{1}~~~
Jinsong Su\textsuperscript{1,2\footnotemark[2]}}\\
\textsuperscript{1}School of Informatics, Xiamen University, China \\
\textsuperscript{2}Key Laboratory of Digital Protection and Intelligent Processing of Intangible Cultural Heritage\\of Fujian and Taiwan (Xiamen University), Ministry of Culture and Tourism, China \\
\texttt{\small{\{litong, zhwang, liangyingshao\}@stu.xmu.edu.cn}}~~~
\texttt{\small{\{xlzheng, xlwang, jssu\}@xmu.edu.cn}}
}

\begin{document}
\maketitle
\begin{abstract}
Recently, the text-to-table generation task has attracted increasing attention due to its wide applications. In this aspect, the dominant model \citep{wu-etal-2022-text-table} formalizes this task as a sequence-to-sequence generation task and serializes each table into a token sequence during training by concatenating all rows in a top-down order. However, it suffers from two serious defects: 1) the predefined order introduces a wrong bias during training, which highly penalizes shifts in the order between rows; 2) the error propagation problem becomes serious when the model outputs a long token sequence. In this paper, we first conduct a preliminary study to demonstrate the generation of most rows is order-insensitive. Furthermore, we propose a novel sequence-to-sequence\&set text-to-table generation model. 
Specifically, 
in addition to a \emph{text encoder} encoding the input text,
our model is equipped with a \emph{table header generator} to first output a table header, i.e., the first row of the table, in the manner of sequence generation. Then we use a \emph{table body generator} with learnable row embeddings and column embeddings to generate a set of table body rows in parallel.
Particularly,
to deal with the issue that there is no correspondence between each generated table body row and target during training, 
we propose a target assignment strategy based on the bipartite matching between the first cells of generated table body rows and targets. Experiment results show that our model significantly surpasses the baselines, achieving state-of-the-art performance on commonly-used datasets.\footnote{We release the code at \url{https://github.com/DeepLearnXMU/seq2seqset}}

\end{abstract}
\renewcommand{\thefootnote}{\fnsymbol{footnote}}
\footnotetext[1]{Equal contribution.}
\footnotetext[2]{Corresponding author.}
\renewcommand{\thefootnote}{\arabic{footnote}}

\begin{figure}[!t]
    \centering
    \includegraphics[width=0.9\linewidth]{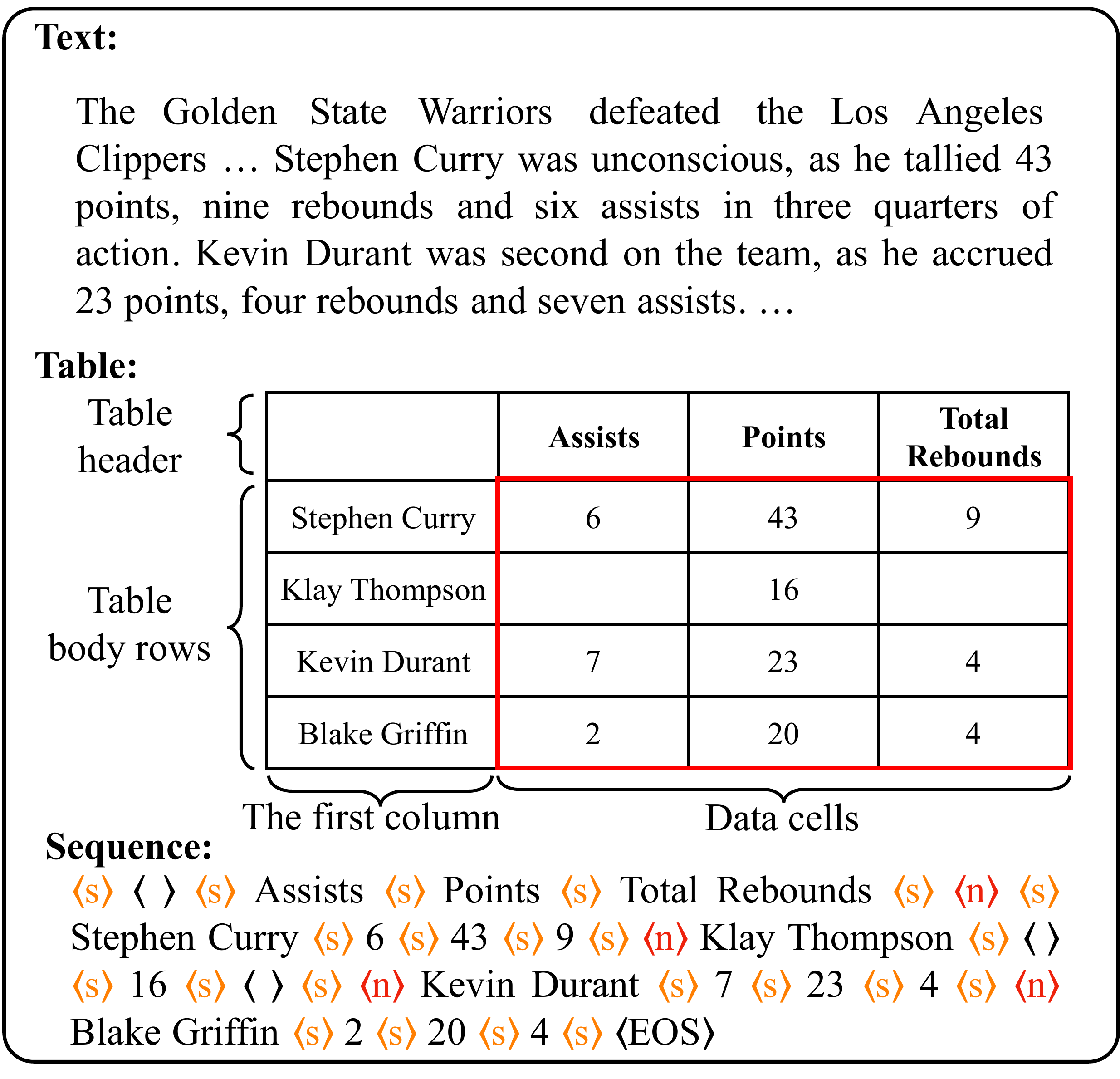}
    \caption{An example of text-to-table task. The input text is a report of a basketball game. In the existing dominant model \citep{wu-etal-2022-text-table}, the output table is serialized into a token sequence during training by concatenating all rows in a top-down order. 
Here, \textcolor{orange}{$\sep$} token is used to separate the cells of each row, \textcolor{red}{$\nline$} token is utilized to separate rows, and $\langle \ \  \rangle$ token means an empty cell. Unlike \citet{wu-etal-2022-text-table}, in this work, we model the generation of each table as a table header and then a set of table body rows. Note that these rows can be further decomposed into the first column and data cells wrapped in the red box.}
    \label{fig:example} 
\end{figure}
\section{Introduction} \label{sec:introduction}
Text-to-table generation is a task that aims to generate a tabular description of important information for a given text. As shown in Figure \ref{fig:example}, the input text is a post-game summary of a basketball game, and the output is a table containing statistics about players. Usually, this task can be widely used to extract important structured information, such as restaurant reviews \citep{novikova-etal-2017-e2e}, team and player statistics \citep{wiseman-etal-2017-challenges}, Wikipedia infoboxes \citep{Bao_Tang_Duan_Yan_Lv_Zhou_Zhao_2018} and biographies \citep{lebret-etal-2016-neural}, benefiting humans understand the input text more intuitively.

In this aspect, \citet{wu-etal-2022-text-table} first propose sequence-to-sequence (seq2seq) text-to-table generation models. They first serialize each table to a token sequence by concatenating all rows in a top-down order. Back to Figure \ref{fig:example}, they represent each table row as a cell sequence, and then represent the entire table by concatenating all rows. Then, they train seq2seq models fine-tuned from BART \citep{lewis-etal-2020-bart}. During inference, the model generates a table in a token-by-token manner and the generated sequence is eventually split by $\sep$ and $\nline$ to obtain the structured table.

Despite some success, the above-mentioned sequence generation manner leads to two defects in the model. \textbf{First}, imposing the above-mentioned predefined order on generating rows in the dataset may bring wrong bias to the model training \citep{ye-etal-2021-one2set, lu-etal-2022-fantastically}. Here, we still take Figure \ref{fig:example} as an example. Each row represents the statistics of a basketball player, and there is no obvious dependency between the statistics of different players. Thus, when considering the order of generating rows, the inconsistent order between generated rows and targets will cause a large training loss, even if the generated rows and targets are exactly the same. \textbf{Second}, as the number of generated rows increases, the outputted token sequence becomes longer, which makes the seq2seq models encounter the serious error propagation problem \citep{ye-etal-2021-one2set, ijcai2021p0542}. Besides, the seq2seq model generates a table autoregressively, of which time complexity is the number of rows times the number of columns, resulting in inefficient GPU acceleration.

In this paper, we first conduct a preliminary study to inspect the effect of row generation order on seq2seq models. Specifically, we randomly reorder table body rows to construct different training datasets. Then, we use these datasets to train seq2seq models, of which performance is compared on the same dataset. Experimental results show that these models exhibit similar performance, proving that the generation of most table body rows is order-insensitive.

Moreover, we propose a novel sequence-to-sequence\&set (Seq2Seq\&set) text-to-table generation model which decomposes the table generation into two steps: generating a table header, i.e., the first row of the table, and then a set of table body rows. As shown in Figure 2, our model mainly consists of three modules: 1) \emph{Text Encoder}. It is a vanilla Transformer encoder, encoding the input document into hidden states; 2) \emph{Table Header Generator} that produces the table header as a token sequence; 3) \emph{Table Body Generator} generating different table body rows in parallel, where each row is generated token by token. To generate different rows from the same text,
we equip the generators with a set of learnable \emph{Row Embeddings}. 
Besides, we add a set of learnable \emph{Column Embeddings} to enhance the semantic consistency between cells in the same column.

During the model training, we need to determine the correspondence between the generated table body rows and targets, so as to achieve the order-independent generation of table body rows. To this end, we propose to use the model to generate the first cells of table body rows. Then, we efficiently determine target assignments according to the matching results between these first cells and those of targets,
which can be modeled as a bipartite matching problem and solved by the Hungarian algorithm \citep{kuhn1955hungarian}. Afterwards, we calculate the training loss based on the one-to-one alignments between the generated table body rows and targets. Besides, during the model inference, we force table body generator to output the same number of cells with the previously-generated generated table header.

Compared with the seq2seq models \citep{wu-etal-2022-text-table}, our model has the following advantages: 1) our model is able to not only alleviate the order bias caused by the sequence generation but also reduce the effect of error propagation on the generation of long sequences; 2) our model achieves faster generation speed since table body rows can be efficiently generated in parallel. 

Experiment results show that our model significantly improves the quality of the generated table, achieving state-of-the-art performance on commonly-used datasets.

\section{Related Work}

Information Extraction (IE) refers to the automatic extraction of structured information such as entities, relationships between entities, and attributes describing entities from unstructured sources \citep{sarawagi2008information}. The common IE tasks include named entity recognition (NER), relation extraction (RE), event extraction (EE), etc.

To achieve high-quality IE, researchers have proposed various task-specific IE methods. With the development of deep learning, researchers mainly focus on neural network based generation models, which are often seq2seq pre-trained models generating serialized structured information. Compared with traditional IE methods, these methods have achieved comparable or even superior results in RE \citep{zeng-etal-2018-extracting, Nayak_Ng_2020}, NER \citep{chen-moschitti-2018-learning, yan-etal-2021-unified-generative}, EE \citep{li-etal-2021-document, lu-etal-2021-text2event}. Along this line, researchers resort to unified models \citep{paolini2021structured, lu-etal-2022-unified} that model multiple IE tasks as the generation of sequences in a uniform format.

In this work, we mainly focus on text-to-table generation that aims to generate structured tables from natural language descriptions. Note that text-to-table generation can be considered as the dual task of table-to-text generation, which intends to generate a textual description conditioned on the input structured data. Usually, these structured data are represented as tables \citep{wiseman-etal-2017-challenges, thomson-etal-2020-sportsett, chen-etal-2020-logical} or sets of table cells \citep{Bao_Tang_Duan_Yan_Lv_Zhou_Zhao_2018, parikh-etal-2020-totto}. Compared with traditional IE tasks, this task does not rely on predefined schemas. In this regard, \citet{wu-etal-2022-text-table} first explore this task as a seq2seq generation task by fine-tuning a BART model \citep{lewis-etal-2020-bart} for generation. By contrast, we model the generation of each table as a table header and then a set of table body rows.
To this end, we propose a Seq2Seq\&set text-to-table model, which can alleviate the defects caused by the sequence generation in the conventional seq2seq models.


\section{Preliminary Study}

We first conduct a preliminary study to inspect the effect of row generation order on the Seq2Seq model \citep{wu-etal-2022-text-table}. Since the table header is the first row of a table containing column names which should be generated first, so we only randomly reorder table body rows to construct different training datasets. Then, we individually train the Seq2Seq models using these datasets with the same setting as the original model, and compare their performance on the same dataset \citep{wiseman-etal-2017-challenges}. 
\begin{table}[!t]
\centering
\small
\tabcolsep=5pt
\begin{tabular}{llccc}
\toprule
\multirow{2}{*}{Subset} & \multirow{2}{*}{Model} & The first  & Table & Data  \\
& & column F1 & header F1 &cell F1 \\
\midrule
\multirow{4}{*}{Team} & Origin & 94.71 & 86.07 & 82.97  \\
& Random1 & 94.57 & 85.93 & 82.83  \\
& Random2 & 94.76 & 86.01 & 83.00  \\
& Random3 & 94.66 & 85.91 & 82.84  \\ 
\midrule 
& $\bm{S}_\mathrm{Team}$ & \textbf{0.08} & \textbf{0.07} & \textbf{0.09}  \\
\midrule
\multirow{4}{*}{Player} & Origin & 92.16 & 87.82 & 81.96 \\
& Random1 & 92.23 & 87.66 & 81.79  \\
& Random2 & 92.33 & 87.69 & 81.84 \\
& Random3 & 92.13 & 87.85 & 81.99 \\
\midrule
& $\bm{S}_\mathrm{Player}$ & \textbf{0.09} & \textbf{0.07} & \textbf{0.10}  \\
\bottomrule
\end{tabular}

\caption{Results of preliminary study on the Team and Player subsets in the Rotowire dataset. We calculate exact match F1 scores on three table parts respectively. ``\emph{Origin}'' means the Seq2Seq model trained using the original dataset, and ``\emph{Random1-3}'' denote models trained on three datasets with different table body row orders, respectively. $S_*$ means the sample standard deviation of the model performance. \protect\footnotemark}
\label{tab:results_random}
\end{table}
\footnotetext{Notice that in \citep{wu-etal-2022-text-table}, they use \emph{row header F1}, \emph{column header F1} and
\emph{non-header cells F1}. Here, we individually rename these to \emph{the first column F1}, \emph{table header F1} and \emph{data cell F1}, so as to avoid ambiguity in descriptions.}

From Table \ref{tab:results_random}, we can observe that the original model and their variants exhibit similar performance.
Besides, we calculate the sample standard deviation of model performance and find that all standard deviations are no more than 0.1.
These results strongly demonstrate that the generation orders of table body rows have a negligible effect on the model performance. In other words, the generation of table body rows is order-insensitive.

\section{Our Model}

\begin{figure*}[!t]
    \centering
    \includegraphics[width=0.9\linewidth]{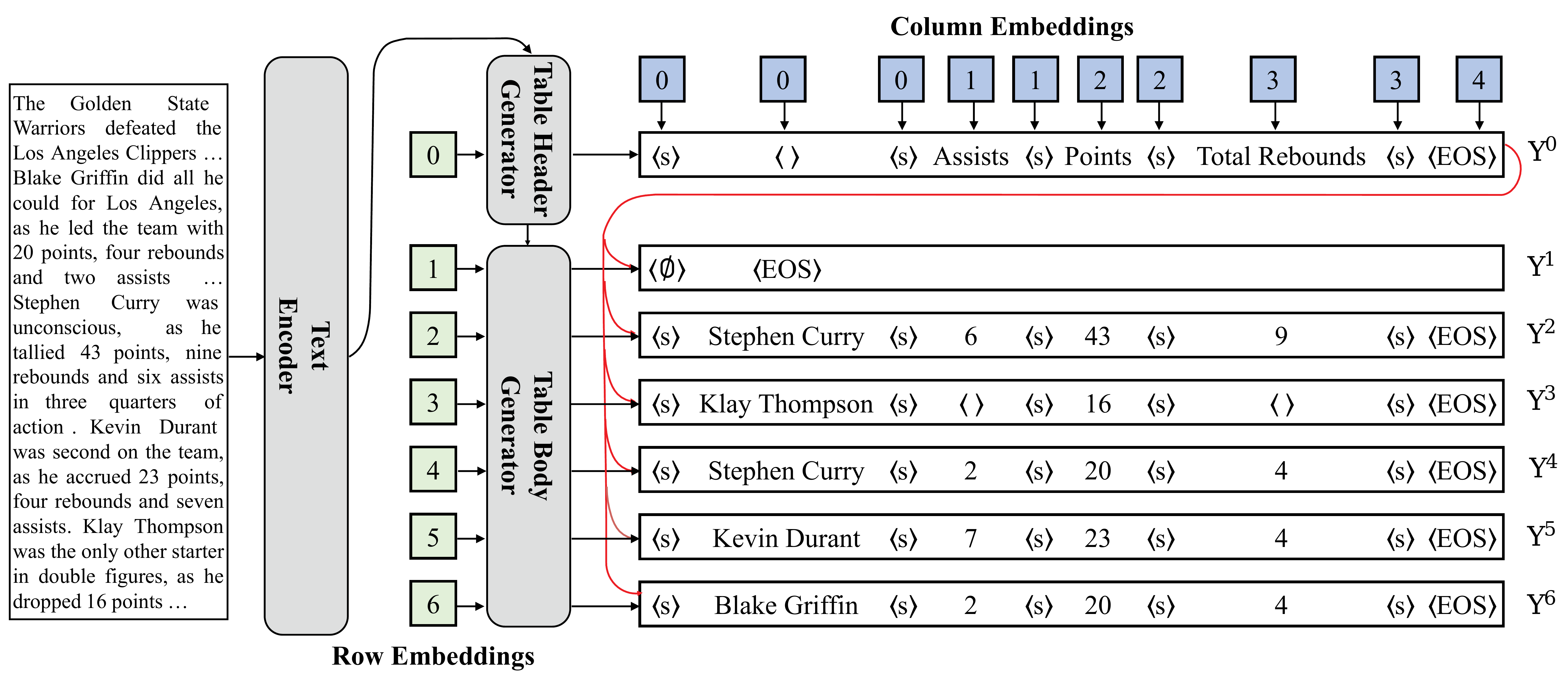}
    \caption{The architecture of our model. It mainly consists of three modules: text encoder encoding input text, table header generator producing a table header as a sequence, and table body generator that generates a set of table body rows in parallel. Note that we have two kinds of learnable embeddings: row embeddings (green squares) enabling table body generator to produce different table body rows, and column embeddings (blue squares) used to enhance the semantic consistency between cells in the same column.}
    \label{fig:model} 
\end{figure*}

In this section, we describe our model in detail. As shown in Figure \ref{fig:model}, our model is composed of three modules: \emph{Text Encoder}, \emph{Table Header Generator} and \emph{Table Body Generator}. Then, we give a detailed description of the model training.

\subsection{Text Encoder} \label{sec:text_encoder}

Our text encoder is used to encode input documents. It is identical to the BART \citep{lewis-etal-2020-bart} encoder, consisting of $L_e$ Transformer encoder layers. The input document is first tokenized into $\mathrm{X}$$=$$x_1, x_2, ..., x_{|\mathrm{X}|}$ using a byte-level Byte-Pair Encoding \citep{Wang_Cho_Gu_2020} tokenizer.

Then, text encoder iteratively updates the hidden states in the following way:
\begin{align}
    \bm{\mathrm{A}}_e^{(l)} &= \mathrm{MultiHead}(\bm{\mathrm{H}}_e^{(l)}, \bm{\mathrm{H}}_e^{(l)}, \bm{\mathrm{H}}_e^{(l)}) ,\\
    \bm{\mathrm{H}}_e^{(l+1)} &= \mathrm{FFN}(\bm{\mathrm{A}}_e^{(l)}) ,
\end{align}
where $\bm{\mathrm{H}}_e^{(l)}$$\in$$\mathbb{R}^{|\mathrm{X}| \times d}$ is the hidden states at the $l$-th layer, and $d$ is the dimension of embeddings and hidden states. $\rm MultiHead(\cdot)$ is a multi-head attention function and $\rm FFN(\cdot)$ refers to a feed-forward network.  We initialize $\bm{\mathrm{H}}_e^{(0)}$ as the sum of $\mathrm{Word}(\mathrm{X})$ and $\mathrm{Pos}(\mathrm{X})$, where $\rm Word(\cdot)$ is a word embedding function and $\rm Pos(\cdot)$ is a learnable position embedding function. Note that we omit the descriptions of layer normalization and residual connection in each sublayer. Please refer to \citep{NIPS2017_3f5ee243, lewis-etal-2020-bart} for more details.

The above process iterates $L_e$ times. Finally, we obtain the hidden states $\bm{\mathrm{H}}_e^{(L_e)}$ of the input document, which will provide useful information for both table header generator and table body generator via attention mechanisms.

\subsection{Table Header Generator} \label{sec:header_generator}

As mentioned previously, our model is designed to generate a table as a table header and a set of table body rows. To this end, we follow previous studies \citep{Zhang_Su_Qin_Liu_Ji_Wang_2018, SU2019103168} to decomposes the table generation into two steps.
We first propose table header generator, which is the same as the BART decoder and generates the table header $\rm Y^0$$=$$y^0_1, y^0_2, ..., y^0_{|\rm Y^0|}$ as a token sequence.

Given the encoder hidden states $\bm{\mathrm{H}}_e^{(L_e)}$, our generator produces the table header in an auto-regressive manner:
\begin{align}
            \bm{\mathrm{A}}_{d0}^{(l)} &= \mathrm{MultiHead}(\bm{\mathrm{H}}_{d0}^{(l)}, \bm{\mathrm{H}}_{d0}^{(l)}, \bm{\mathrm{H}}_{d0}^{(l)}), \\
    \bm{\mathrm{\overline{A}}}_{d0}^{(l)} &= \mathrm{MultiHead}(\bm{\mathrm{A}}_{d0}^{(l)}, \bm{\mathrm{H}}_e^{(L_e)}, \bm{\mathrm{H}}_e^{(L_e)}) ,\\
    \bm{\mathrm{H}}_{d0}^{(l+1)} &= \mathrm{FFN}(\bm{\mathrm{\overline{A}}}_{d0}^{(l)}) , 
\end{align}
where $\bm{\mathrm{H}}_{d0}^{(l)} \in \mathbb{R}^{|\mathrm{Y^0}|\times d}$ is the hidden states at the $l$-th layer. In the first layer, we initialize the $j$-th decoder input as the sum of $\mathrm{Word}(y^0_{j-1})$, $\mathrm{Pos}(y^0_{j-1})$, $\mathrm{Row}(y^0_{j-1})$ and $\mathrm{Col}(y^0_{j-1})$, where $\rm Word(\cdot)$ and $\rm Pos(\cdot)$ share parameters with text encoder, row embeddings $\rm Row(\cdot)$  and column embeddings $\rm Col(\cdot)$ will be described in the next subsection.

Finally, after iterating the above process for $L_d$ times, we obtain the last-layer hidden states $\bm{\mathrm{H}}_{d0}^{(L_d)} = \{\mathrm{H}^{(L_d)}_{d0,j}\}_{1\leq j \leq |\mathrm{Y}^0|}$, where $\mathrm{H}_{d0,j}^{(L_d)}$ is used to output the $j$-th target token:
\begin{equation}
    y^0_j = \mathrm{argmax}(\bm{\mathrm{W}}_o \mathrm{H}_{d0,j}^{(L_d)}) ,
\end{equation}
where $\bm{\mathrm{W}}_o \in \mathbb{R}^{\rm|V|\times d}$ is a learnable parameter matrix, and $\rm V$ is the target vocabulary. 
 
\subsection{Table Body Generator} \label{sec:body_generator}

To generate a set of table body rows in parallel, we propose a novel semi-autoregression table body generator. It is also stacked with $L_d$ layers, each of which consists of self-attention, cross-attention and feed-forward network sublayers. Particularly, it shares parameters with table header generator, so that it can directly exploit the hidden states of table header generator via self-attention. This generator produces $M$ table body rows \{$\mathrm{Y}^m\}_{1\leq m \leq M}$ in parallel, under the semantic guidence of $\bm{\mathrm{H}}_e^{(L_e)}$ and $\{\bm{\mathrm{H}}_{d0}^{(l)}\}_{1\leq l \leq L_d}$. Particularly, we use a special $\nullrow$ token to represent ``no corresponding row''. Formally, the $m$-th table body row, $\mathrm{Y}^m = y^m_1, y^m_2, ..., y^m_{|\mathrm{Y}^m|}$ is generated in an auto-regressive way:
\begin{align}
    \bm{\mathrm{\overline{H}}}_{dm}^{(l)} &= [\bm{\mathrm{H}}_{d0}^{(l)}; \bm{\mathrm{H}}_{dm}^{(l)}] , \\
    \bm{\mathrm{A}}_{dm}^{(l)} &= \mathrm{MultiHead}(\bm{\mathrm{H}}_{dm}^{(l)}, \bm{\mathrm{\overline{H}}}_{dm}^{(l)}, \bm{\mathrm{\overline{H}}}_{dm}^{(l)}) ,\\
    \bm{\mathrm{\overline{A}}}_{dm}^{(l)} &= \mathrm{MultiHead}(\bm{\mathrm{A}}_{dm}^{(l)}, \bm{\mathrm{H}}_e^{(L_e)}, \bm{\mathrm{H}}_e^{(L_e)}) ,\\
    \bm{\mathrm{H}}_{dm}^{(l+1)} &= \mathrm{FFN}(\bm{\mathrm{\overline{A}}}_{dm}^{(l)}) , 
\end{align}
where $\bm{\mathrm{H}}_{dm}^{(l)}$ $\in$ $\mathbb{R}^{|\mathrm{Y}^m|\times d}$ is the hidden states for generating $\mathrm{Y}^m$. Note that our generator exploits both hidden states of table header and previous tokens in the same row to produce a table body row. 

Taking the sum of word embeddings and positional embeddings as inputs, the vanilla Transformer decoder can only generate a sequence but not a set. To generate a set of table body rows in parallel, we introduce $M$ additional learnable embeddings called \emph{Row Embeddings} (See the green squares in Figure \ref{fig:model}) into inputs, guiding table body generator to produce different rows. Here, $M$ is a predefined parameter that is usually larger than the maximum number of rows in training data.  Furthermore, to enhance the semantic consistency between cells in the same column, we add another learnable embeddings named \emph{Column Embeddings} (See the blue squares in Figure \ref{fig:model}) into inputs. Column embeddings are similar to positional embeddings but are defined at the cell level. By doing so, tokens of cells in the same column are equipped with identical column embeddings.

Formally, with row and column embeddings, the initial hidden state of our generator becomes
\begin{equation}
    \begin{aligned}
    \mathrm{H}_{dm,k}^{(0)} &= \mathrm{Word}(y^m_{k-1}) + \mathrm{Pos}(y^m_{k-1})     \\ 
    &+ \mathrm{Row}(y^m_{k-1}) + \mathrm{Col}(y^m_{k-1}),
    \end{aligned}
\end{equation}
where $y^m_{k-1}$ is the ($k$$-$$1$)-th output token at the $m$-th table body row, and $\rm Row(\cdot)$ and $\rm Col(\cdot)$ are row and column embedding functions, respectively. 

Through $L_d$ times of hidden state updates, we obtain the last-layer hidden states \{$\bm{\mathrm{H}}_{dm}^{(L_d)}\}_{1\leq m \leq M}$. Finally, based on $\mathrm{H}^{(L_d)}_{dm,k}$, we obtain the token with maximum probability as the output:
\begin{equation}
    y^m_k = \mathrm{argmax}(\bm{\mathrm{W}}_o \mathrm{H}_{dm,k}^{(L_d)}).
\end{equation}

In order to maintain an equal number of cells in every row, table body generator keeps generating a row until the number of $\sep$ matches that in the header.

\subsection{Training} \label{sec:training}

\textbf{Training Loss} As mentioned above, we decompose the generation of a table into two steps. Correspondingly, we define the training loss as:
\begin{equation}
    \mathcal{L} = \lambda \mathcal{L}_h + (1-\lambda) \mathcal{L}_{b},
\end{equation}
where $\lambda$ is a hyper-parameter to balance the effect of \emph{the table header generation loss} $\mathcal{L}_h$ and \emph{the table body generation loss} $\mathcal{L}_{b}$. 

As for $\mathcal{L}_h$, we follow common practice to define $\mathcal{L}_h$ as a cross-entropy loss between the predictive distributions of the generated table header and the target one:
\begin{equation}
    \mathcal{L}_h = -\sum_{j=1}^{|\mathrm{Y}^0|}\mathrm{log}\hat{p}^0_j(y^0_j),
\end{equation}
where $\hat{p}^0_j(\cdot)$ is the predictive probability of the $j$-th token in the table header $\mathrm{Y}^0$ using teacher forcing.

\textbf{Target Assignments Based on the First Cells} We also define $\mathcal{L}_b$ as a cross-entropy loss between the predictive distributions of the generated table body rows and targets. However, there is no correspondence between each generated table body row and target during training, and hence we can not directly calculate $\mathcal{L}_b$. To deal with this issue, we learn from the recent studies on set generation \citep{10.1007/978-3-030-58452-8_13, ye-etal-2021-one2set, xie-etal-2022-wr} and propose to efficiently determine target row assignments according to the matching results between the first cells of generated table body rows and those of targets. Here, we use the first cell to represent the whole row, because it is usually unique and often contains the important information of a table such as a name and a primary key.

Concretely, we first use our model to generate the first cells of all table body rows. During this process, we obtain generation probability distributions $\{\bm{\mathrm{P}}^{m}\}_{1\leq m \leq M}$, where $\bm{\mathrm{P}}^{m} = \{p^m_k\}_{1\leq k \leq |\bm{\mathrm{P}}^{m}|}$ and $p^m_k$ is the predictive distribution at the $k$-th timestep for the $m$-th table body row. Particularly, we pad the set of target table body rows to size $M$ with $\nullrow$. Afterwards, we determine the target assignments via the bipartite matching between the generated table body rows and targets: 
\begin{equation}
    \hat{f} = \mathop{\mathrm{argmin}}\limits_{f\in \mathrm{F}(M)}\sum_{m=1}^M \mathcal{C}(\mathrm{Y}^{f(m)}, \bm{\mathrm{P}}^{m}),
\end{equation}
where $\mathrm{F}(M)$ denotes the set of all $M!$ one-to-one mapping functions and $f(\cdot)$ is a function aligning the $m$-th generated table body row to the $f(m)$-th target one. The optimal matching can be efficiently determined with the Hungarian algorithm \citep{kuhn1955hungarian}. More specifically, the matching cost $\mathcal{C(\cdot)}$ takes into account the token level probability, which is defined as follows:
\begin{equation}
    \mathcal{C}(\mathrm{Y}^{f(m)},\bm{\mathrm{P}}^m) = -\sum_{k=1}^{N} \mathbbm{1}_{\{\mathrm{Y}\neq \varnothing \}}p^m_k(y^{f(m)}_k),
\end{equation}
where $N$ is the length of the first cell in $\mathrm{Y}^{f(m)}$ and $p^m_k(y^{f(m)}_k)$ is the predictive probability of the $k$-th target token $y^{f(m)}_k$ of the $m$-th table body row. We ignore the score from matching predictions with $\nullrow$, so as to ensure that each generated row can be aligned with a non-empty target as possible.
\begin{figure}[tb]
    \centering
    \includegraphics[width=0.9\linewidth]{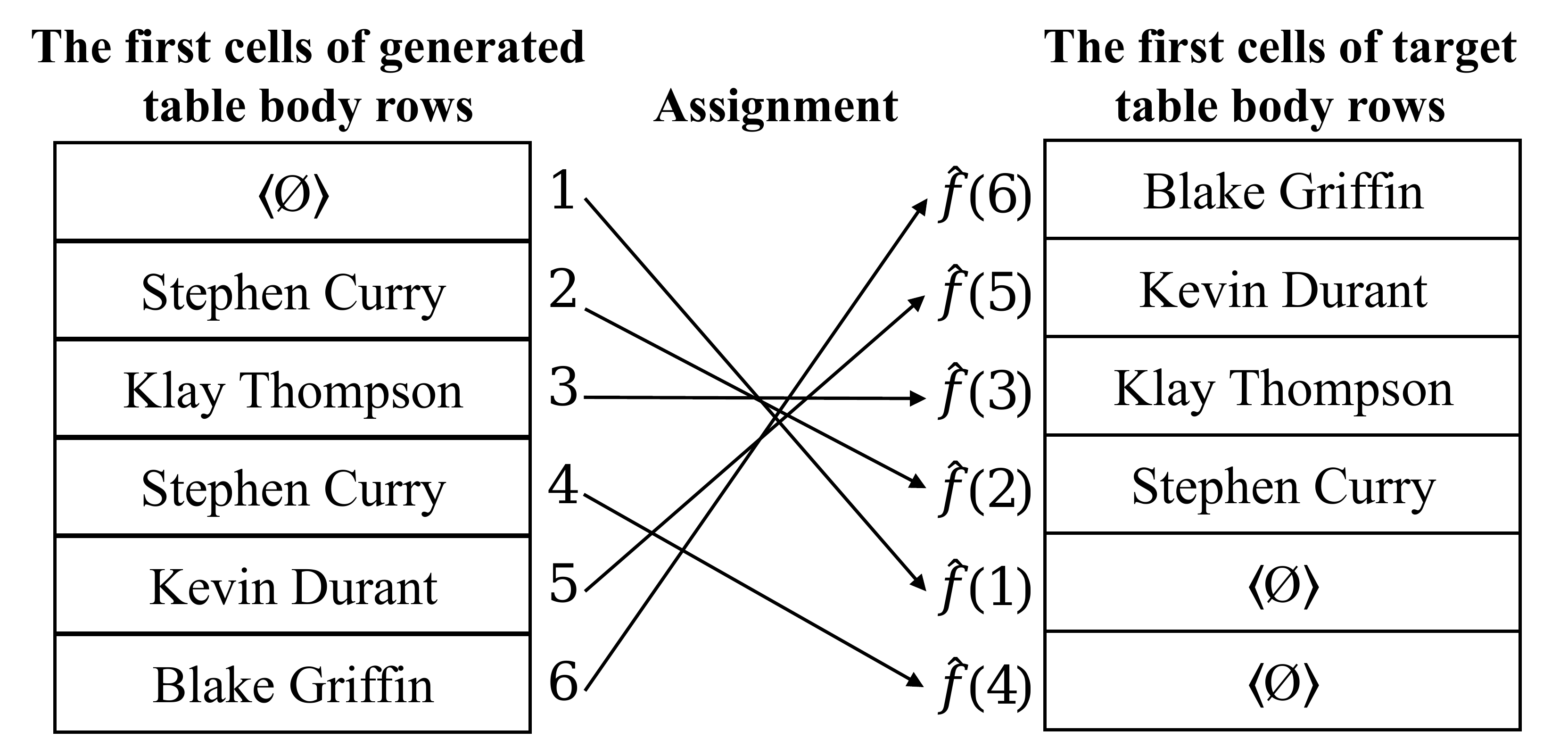}
    \caption{The bipartite matching between generated table body rows and targets. We use the first cell to represent each table body row, and determine target assignments according to the optimal bipartite matching between the generated first cells and those of targets. The function $\hat{f}(\cdot)$ maps the index of the input generated table body row to the corresponding target one.}
    \label{fig:match} 
\end{figure}

For example, in Figure \ref{fig:match}, our model generates the first cells of six table body rows, where the first one is $\nullrow$ token and the others are player names. Then we assign each generated table body row to a target one according to the above-mentioned bipartite matching. In this example, we can find an optimal matching, with a mapping function $\hat{f}$ satisfying that
$\hat{f}(1) = 5, \hat{f}(2) = 4, ..., \hat{f}(6) = 1$. Note that there are two ``\emph{Stephen Curry}'' occurring in row 2 and row 4, but are aligned to different targets due to the one-to-one matching. In this way, we can guarantee that supervision signal for generating each table body row is unique, alleviating the generation of duplicate rows.

Finally, through the above target row assignments, we can calculate $\mathcal{L}_b$ as follows:
\begin{equation}
    \mathcal{L}_b = -\sum_{m=1}^M\sum_{k=1}^{|\mathrm{Y}^{\hat{f}(m)}|}\mathrm{log}\hat{p}^m_k(y^{\hat{f}(m)}_k),
\end{equation}
where $\hat{p}^m_k(\cdot)$ is the predictive distribution of the $m$-th generated table body row at the timestep $k$, and $y^{\hat{f}(m)}_k$ is the $k$-th token in the assigned target row.

Particularly, inspired by the recent studies \citep{10.1007/978-3-030-58452-8_13, ye-etal-2021-one2set, xie-etal-2022-wr}, when $y_k^{\hat{f}(m)}$ $=$ $\nullrow$, we multiply its token-level loss with a predefined factor to down-weight its effect, so as to reduce the negative effect of excessive $\nullrow$ tokens.

\section{Experiments}

\subsection{Setup}
\textbf{Datasets} 
Following the previous work \citep{wu-etal-2022-text-table}, we conduct experiment on four commonly-used datasets for table-to-text generation: Rotowire \citep{wiseman-etal-2017-challenges}, E2E \citep{novikova-etal-2017-e2e}, WikiTableText \citep{Bao_Tang_Duan_Yan_Lv_Zhou_Zhao_2018}, and WikiBio \citep{lebret-etal-2016-neural}. As the main dataset of our experiments, Rotowire has two types of tables named Team and Player. In Rotowire, each instance has multiple columns while the other three datasets have only two columns. We use the processed datasets from \citep{wu-etal-2022-text-table}.

\begin{table*}[!htb]
\centering
\small
\begin{tabular}{l|l|ccc|ccc|ccc|c}
\toprule
\multirow{2}{*}{Subset}& \multirow{2}{*}{Model} & \multicolumn{3}{c|}{The first column F1 } & \multicolumn{3}{c|}{Table header F1 } & \multicolumn{3}{c|}{Data cell F1 } & Error \\
& & Exact & Chrf & BERT & Exact & Chrf & BERT & Exact & Chrf & BERT & rate \\
\midrule
\multirow{5}{*}{Team} & Sent-level RE & 85.28 & 87.12 & 93.65 & 85.54 & 87.99 & 87.53 & 77.17 & 79.10 & 87.48 & 0.00 \\
& Doc-level RE & 84.90 & 86.73 & 93.44 & 85.46 & 88.09 & 87.99 & 75.66 & 77.89 & 87.82 & 0.00 \\
& Seq2Seq & 94.71 & 94.93 & 97.35
 & \textbf{86.07} & 89.18 & 88.90 & 82.97 & 84.43 & 90.62 & 0.49 \\
& Seq2Seq-c & 94.97 & 95.20 & 97.51 & 86.02 & 89.24 & 89.05 & 83.36 & 84.76 & 90.80 & 0.00 \\
& Seq2Seq\&set & \textbf{96.80}$^\ddagger$ & \textbf{97.10}$^\ddagger$ & \textbf{98.45}$^\ddagger$ & 86.00 & \textbf{89.48} & \textbf{93.11}$^\ddagger$ & \textbf{84.33}$^\ddagger$ & \textbf{85.68}$^\ddagger$ & \textbf{91.30}$^\ddagger$ & 0.00 \\
\midrule
\multirow{5}{*}{Player} & Sent-level RE & 89.05 & 93.00 & 90.98 & 86.36 & 89.38 & 93.07 & 79.59 & 83.42 & 85.35 & 0.00 \\
& Doc-level RE & 89.26 & 93.28 & 91.19 & 87.35 & 90.22 & 97.30 & 80.76 & 84.64 & 86.50 & 0.00 \\
& Seq2Seq & 92.16 & 93.89 & 93.60 & 87.82 & 91.28 & 94.44 & 81.96 & 84.19 & 88.66 & 7.40 \\
& Seq2Seq-c & 92.31 & 94.00 & 93.71 & 87.78 & 91.26 & 94.41 & 82.53 & 84.74 & 88.97 & 0.00 \\
& Seq2Seq\&set & \textbf{92.83}$^\dagger$ & \textbf{94.48}$^\dagger$ & \textbf{96.43}$^\ddagger$ & \textbf{88.02} & \textbf{91.60}$^\dagger$ & \textbf{95.08}$^\dagger$ & \textbf{83.51}$^\ddagger$ & \textbf{85.75}$^\ddagger$ & \textbf{90.93}$^\ddagger$ & 0.00 \\
\bottomrule
\end{tabular}
\caption{Results of baselines and our Seq2Seq\&set model on Rotowire. We show the F1 score based on exact match (Exact), chrf score (Chrf), and BERTScore (BERT).  $\dagger$/$\ddagger$ indicates significant at $p$$<$$0.05$/$0.01$ over Seq2Seq-c with 1000 bootstrap tests \citep{efron1994introduction}.}
\label{tab:results_rotowire}
\end{table*}

\textbf{Implementation Details}
We initialize our model with the pre-trained BART-base \citep{lewis-etal-2020-bart}, which consists of 6 encoder layers and 6 decoder layers. The number of multi-head attention is 12, the dimension of embedding and hidden state is 768, and the dimension of feed-forward network is 3,072. We reuse the vocabulary from the pre-trained BART-base model, whose size is 51,200. We use the Adam \citep{DBLP:journals/corr/KingmaB14} optimization algorithm with a fixed maximum number of tokens as 4,096. For different datasets, we set different numbers of row embeddings according to the maximum row numbers in training sets. We train a separate model on each dataset and select the model with the lowest validation loss. 

\textbf{Baselines}
We compare our model with the following baselines mentioned in \citep{wu-etal-2022-text-table}:
\begin{itemize}[itemsep=2pt,topsep=0pt,parsep=0pt]
    \item \textbf{Sent-level RE} This model uses an existing method of relation extraction (RE) \citep{zhong-chen-2021-frustratingly} to extract information based on predefined schemas. It takes the first column and data cells as entities and the types of table header cells as relations.
    \item \textbf{Doc-level RE} It applies the same RE method, except that it predicts the relations between entities within multiple sentences. 
    \item \textbf{NER} It is a BERT-based \citep{devlin-etal-2019-bert} entity extraction method that considers data cells in each table as entities and its first column cells as entity types.
    \item \textbf{Seq2Seq} \citep{wu-etal-2022-text-table} It is a Transformer based seq2seq model that models the generation of a table as a sequence.
    \item \textbf{Seq2Seq-c} \citep{wu-etal-2022-text-table} This model is a Seq2Seq variant, where the cell number of each table body row is limited to the same as that of table header.
\end{itemize}

\textbf{Evaluation}
We use the same evaluation script from \citep{wu-etal-2022-text-table}. We adopt the F1 score as the evaluation measure, 
which is calculated in the following way: the precision and recall are first computed to get table-specific F1 scores, which are then averaged to obtain the final score. 
Here, precision is defined as the percentage of correctly predicted cells among the generated cells, and recall is defined as the percentage of correctly predicted cells among target cells. Particularly, the F1 score is calculated in three ways: \emph{exact match} that matches two cells exactly, \emph{chrf score} that calculates character-level n-gram similarity, and \emph{BERTscore} that calculates the similarity between BERT embeddings of two cells. 

For Rotowire, we report the F1 scores of the first column, table header and data cells, which refer to \emph{row header F1}, \emph{column header F1} and \emph{non-header cells F1} in \citep{wu-etal-2022-text-table}. For the other three datasets, there are only fixed two columns, so the F1 score of table header is not calculated. For data cells, we use not only the content but also the table header/the first column cells to ensure that the cell is on the right column/row. Note that these metrics are insensitive to the orders of rows and columns. Besides, we calculate the error rate to represent the percentage of erroneous format tables.

\begin{table}[!t]
\centering
\tabcolsep=4pt
\small
\begin{tabular}{lccc}
\toprule
\multirow{2}{*}{Model} & \multicolumn{3}{c}{\# Sentences per second (speedup)} \\ \cmidrule{2-4}
& Team  & Player & E2E  \\
\midrule 
Seq2Seq & 1.24 (1.00$\times$) & 0.32 (1.00$\times$) & 1.66 (1.00$\times$) \\
Seq2Seq-c & 1.22 (0.98$\times$) & 0.30 (0.94$\times$) & 1.62 (0.98$\times$)  \\
Seq2Seq\&set & 1.84 (1.48$\times$) & 1.09 (3.41$\times$) & 5.96 (3.59$\times$)  \\
\bottomrule
\end{tabular}
\caption{Inference efficiency comparison among different models. Here, we conduct experiments on an NVIDIA RTX 3090 GPU.}
\label{tab:results_speed}
\end{table}

\subsection{Main Results}
\begin{table*}[!t]
\centering
\small
\begin{tabular}{l|l|ccc|ccc|c}
\toprule
\multirow{2}{*}{Dataset} & \multirow{2}{*}{Model} & \multicolumn{3}{c|}{The first column F1} & \multicolumn{3}{c|}{Data cell F1} & Error \\
& & Exact & Chrf & BERT & Exact & Chrf & BERT & rate \\
\midrule
\multirow{4}{*}{E2E} & NER & 91.23 & 92.40 & 95.34 & 90.80 & 90.97 & 92.20 & 0.00 \\
& Seq2Seq & 99.62 & 99.69 & 99.88 & 97.87 & 97.99 & 98.56 & 0.00 \\
& Seq2Seq-c & \textbf{99.63} & \textbf{99.69} & \textbf{99.88} & 97.88 & 98.00 & 98.57 & 0.00 \\
& Seq2Seq\&set & 99.62 & \textbf{99.69} & 99.83 & \textbf{98.65}$^\ddagger$ & \textbf{98.70}$^\ddagger$ & \textbf{99.08}$^\ddagger$ & 0.00 \\
\midrule
\multirow{4}{*}{WikiTableText} & NER & 59.72 & 70.98 & 94.36 & 52.23 & 59.62 & 73.40 & 0.00 \\
& Seq2Seq & 78.15 & 84.00 & 95.60 & 59.26 & 69.12 & 80.69 & 0.41 \\
& Seq2Seq-c & 78.16 & 83.96 & 95.68 & 59.14 & 68.95 & 80.74 & 0.00 \\
& Seq2Seq\&set & \textbf{78.67}$^\ddagger$ & \textbf{84.21}$^\ddagger$ & \textbf{95.88} & \textbf{59.94}$^\ddagger$ & \textbf{69.59}$^\ddagger$ & \textbf{81.67}$^\ddagger$ & 0.00 \\
\midrule
\multirow{4}{*}{WikiBio} & NER & 63.99 & 71.19 & 81.03 & 56.51 & 62.52 & 61.95 & 0.00 \\
& Seq2Seq & 80.53 & 84.98 & 92.61 & 68.98 & 77.16 & 76.54 & 0.00 \\
& Seq2Seq-c & 80.52 & 84.96 & 92.60 & 69.02 & 77.16 & 76.56 & 0.00 \\
& Seq2Seq\&set & \textbf{81.03}$^\dagger$ & \textbf{85.44}$^\dagger$ & \textbf{93.02}$^\dagger$ & \textbf{69.51}$^\ddagger$ & \textbf{77.53}$^\ddagger$ & \textbf{77.13}$^\ddagger$ & 0.00 \\
\bottomrule
\end{tabular}
\caption{Results of baselines and our Seq2Seq\&set model on E2E, WikiTableText and WikiBio.}
\label{tab:results_wiki}
\end{table*}

Table \ref{tab:results_rotowire} reports the results on the Team and Player subsets of Rotowire.  We observe that our model consistently outperforms all baselines in terms of three kinds of F1 scores. Particularly, in terms of data cell F1, which is the most difficult of the three kinds of F1 scores, ours achieves significant improvements. Besides, note that both Seq2Seq-c and our model enforce the number of cells in each table body row to be the same as that of table header, so their error rates are $0$.
Table \ref{tab:results_wiki} shows the results on E2E, WikiTableText and WikiBio. Likewise, our model outperforms almost all baselines.

\subsection{Inference Efficiency}

We compare the inference efficiency of different models.
From Table \ref{tab:results_speed}, we observe that ours is significantly more efficient than baselines, due to its advantage in the parallel generation of table body rows.

\begin{figure}[!t]
    \centering
    \includegraphics[width=0.9\linewidth]{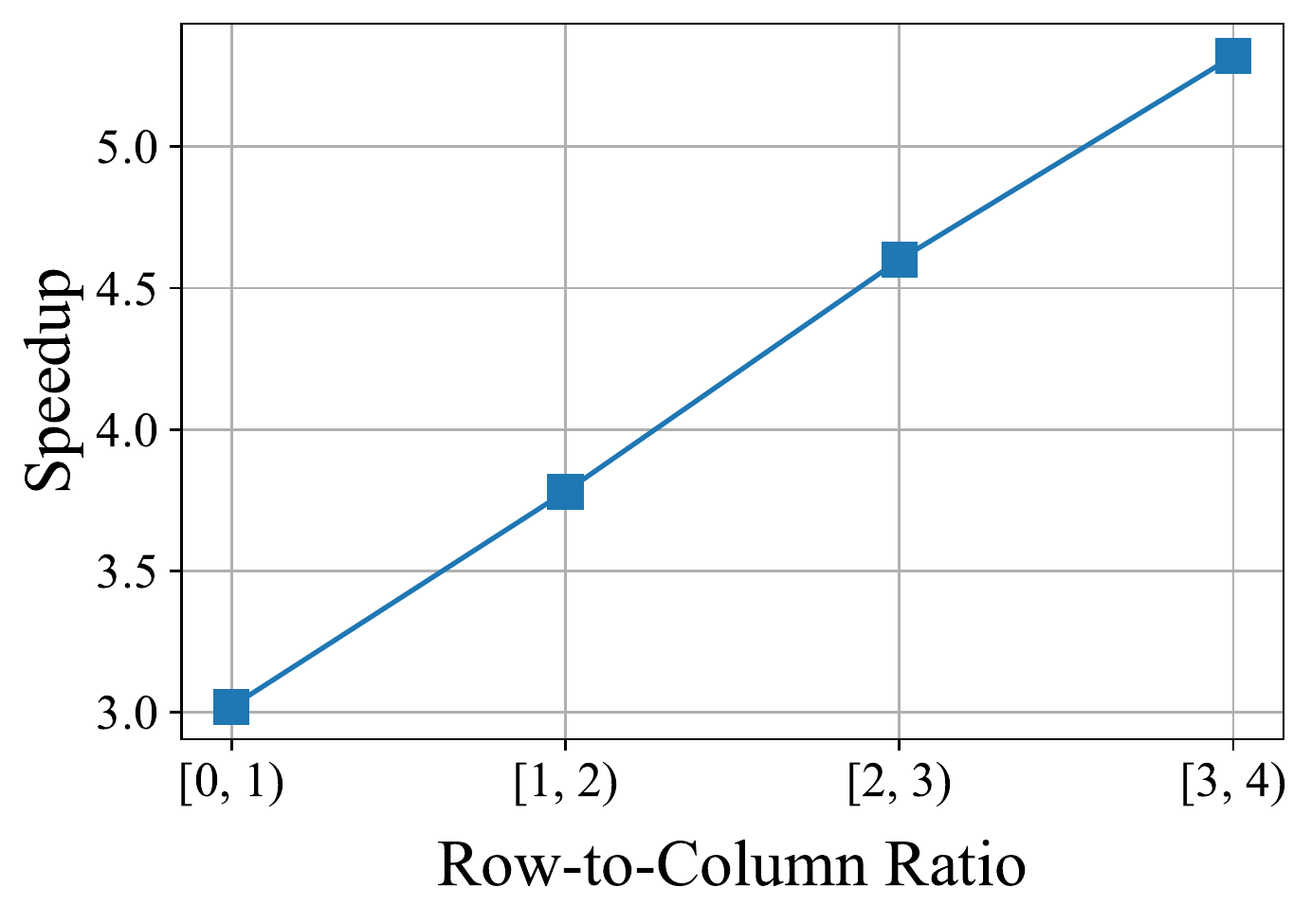}
    \caption{Speedup of Seq2Seq-c to Seq2Seq\&set with different row-to-column ratios.}
    \label{fig:analyze-speed} 
\end{figure}

To investigate the effect of speedup on different types of tables, we carry out experiments on the Rotowire Player dataset. We define \emph{row-to-column ratio} as the row number divided by the column number and measure the speedup with different row-to-column ratios. The results depicted in Figure \ref{fig:analyze-speed} demonstrate that our model exhibits a linear improvement as the row-to-column ratio increases.

\subsection{Error Propagation Analysis}

\begin{figure}[!t]
    \centering
    \includegraphics[width=0.9\linewidth]{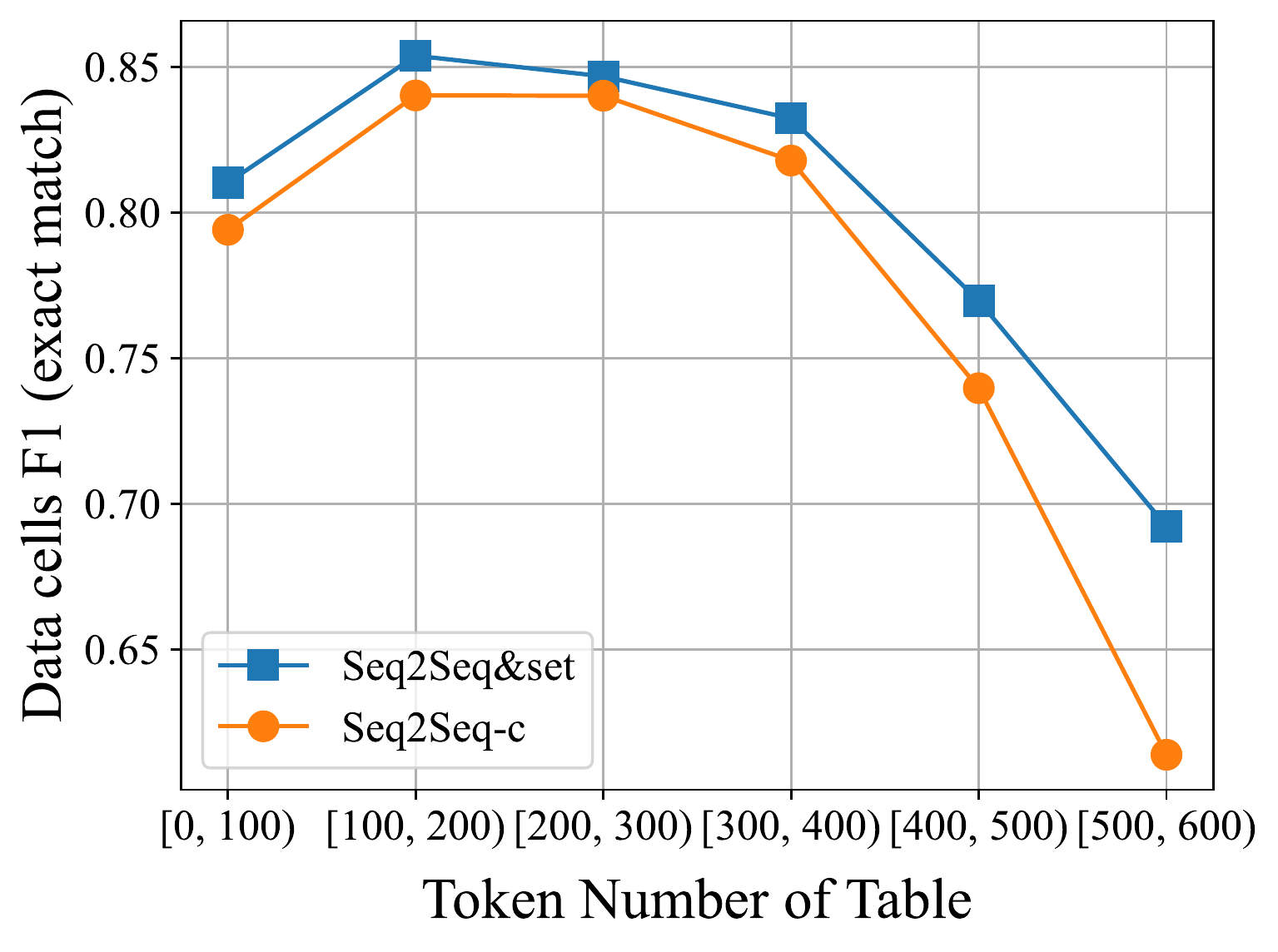}
    \caption{Performance comparison between Seq2Seq-c and Seq2Seq\&set on the tables with different numbers of tokens.}
    \label{fig:analyze} 
\end{figure}

As analyzed in Introduction, our model is able to better alleviate the error propagation issue than previous models.
To verify this, we conduct experiments on the longest dataset Rotowire Player, and then compare the performance of our model and Seq2Seq-c.
From Figure \ref{fig:analyze}, we observe that ours always outperforms the baseline model. Particularly, with the number of table tokens increasing, our model exhibits more significant performance advantages over Seq2Seq-c.

\subsection{Ablation Study}
\begin{table}[!t]
\centering
\small
\tabcolsep=2pt
\begin{tabular}{llccc}
\toprule
 \multirow{2}{*}{Subset}&\multicolumn{1}{c}{\multirow{2}{*}{Model}} &The first & Table  & Data \\
 && column F1 & header F1 & cell F1 \\
 \midrule
 \multirow{5}{*}{Team} 
& Seq2Seq\&set & 96.80 & 86.00 & 84.33 \\
 & \quad w/o row embed. & 66.91 & 85.73 & 57.25 \\
 & \quad w/o col. embed. & 96.75 & 86.04 & 83.32 \\
 & \quad w/o tgt. assign. & 92.87 & 85.87 & 79.43  \\
&  \quad w/o header gen. & 95.34 & 85.23 & 59.02 \\
\midrule
 \multirow{5}{*}{Player}
 &Seq2Seq\&set & 92.83 & 88.02 & 83.51 \\
&\quad w/o row embed. & 31.13 & 87.83 & 26.20 \\
 &\quad w/o col. embed. & 90.06 & 87.47 & 80.18 \\
 &\quad w/o tgt. assign. & 67.71 & 87.99 & 60.22 \\
&\quad w/o header gen. & 92.45 & 86.83 & 58.39 \\
\bottomrule
\end{tabular}
\caption{Results of ablation study. Here, the F1 scores are calculated in the way of exact match.}
\label{tab:ablation}
\end{table}

We conduct an ablation study on Rotowire to verify the effectiveness of different components of our model. The results are shown in Table \ref{tab:ablation}, involving four variants:

\emph{w/o Row Embeddings.}
We remove row embeddings in this variant. From lines 3 and 8, we observe that this variant is completely collapsed. This is reasonable that without row embeddings, the row-specific inputs of table body generator become exactly the same, resulting in the generator failing to generate distinct rows.

\emph{w/o Column Embeddings.}
We remove column embeddings in this variant. From lines 4 and 9, we observe that the data cell F1 decreases a lot. Thus, we also confirm that column embeddings are indeed useful in enhancing the semantic consistency between cells in the same column. The other two scores changed very little, which we believe is due to the fact that table headers and the first columns have no need to refer to the other rows.

\emph{w/o Target Assignments.}
In this variant, we discard the target assignments based on the first cells during training, which makes our model learn to generate table body rows in the original order of targets. As shown in lines 5 and 10, our model exhibits a significant performance drop. 

\emph{w/o Table Header Generator.}
In this variant, we simultaneously generate table header and table body rows in parallel. Consequently, the variant can not leverage the information of generated table header during the process of generating table body rows, and thus exhibits worse performance than the original model. This result proves that it is reasonable to distinguish the generation of table header and table body rows.

\section{Conclusion}
In this paper, we propose a Seq2Seq\&set model for text-to-table generation, which first outputs a table header, and then table body rows. Most importantly, unlike the previous study \cite{wu-etal-2022-text-table}, we model the generation of table body rows as a set generation task, which is able to alleviate not only the wrong bias caused by the predefined order during training, but also the problem of error propagation during inference. Experimental results show that our model gains significant performance improvements over the existing SOTA.

\section*{Limitations}

Our model is currently suitable for generating ordinary tables with attribute names and records, but it may struggle with more complex table formats that involve merged cells. To improve the flexibility of our model, we plan to investigate more versatile forms of table representation.

Another limitation of our model is that our model training involves longer training time, compared with seq2seq baselines. This may be due to the inherent instability of target assignments. In the future, we will explore refining the model training by reducing the target assignment instability.

The existing datasets for this task is relatively simple, and in the future we will conduct experiments on more complex datasets that require reasoning, such as WebNLG \citep{gardent-etal-2017-webnlg}.
\section*{Ethics Statement}
This paper proposes a Seq2Seq\&set model for text-to-table generation. We take ethical considerations seriously and ensure that the research and methods used in this study are conducted in an ethical and responsible manner. The datasets used in this paper are publicly available and have been widely adopted by researchers for testing the performance of text-to-table generation. This study does not involve any data collection or release, and thus there exist no privacy issues. We also take steps to ensure that the findings and conclusions of this study are reported accurately and objectively.

\section*{Acknowledgement}
The project was supported by National Natural Science Foundation of China (No. 62276219), Natural Science Foundation of Fujian Province of China (No. 2020J06001), Youth Innovation Fund of Xiamen (No. 3502Z20206059). We also thank the reviewers for their insightful comments.

\bibliography{acl2023}
\bibliographystyle{acl_natbib}
\appendix

\section{Dataset Statistics} \label{app:statistics}
Table \ref{tab:dataset} shows the statistics of the datasets we used. We list the numbers of instances in training, validation, and test sets and the average number of BPE tokens per instance. We also give the average numbers of rows and columns per instance.

\section{Hyper-parameter Settings} \label{app:hyper}
Table \ref{tab:hyper_parameter} shows the hyper-parameter settings in our experiments. We set the hyper-parameters by referring to the existing work and choosing values that result in the best performance (measured in data cell F1) on the validation sets.

\section{Case Study} \label{app:case}

\begin{wrapfigure}{r}{0.48\textwidth}
    \centering
    \includegraphics[width=\linewidth]{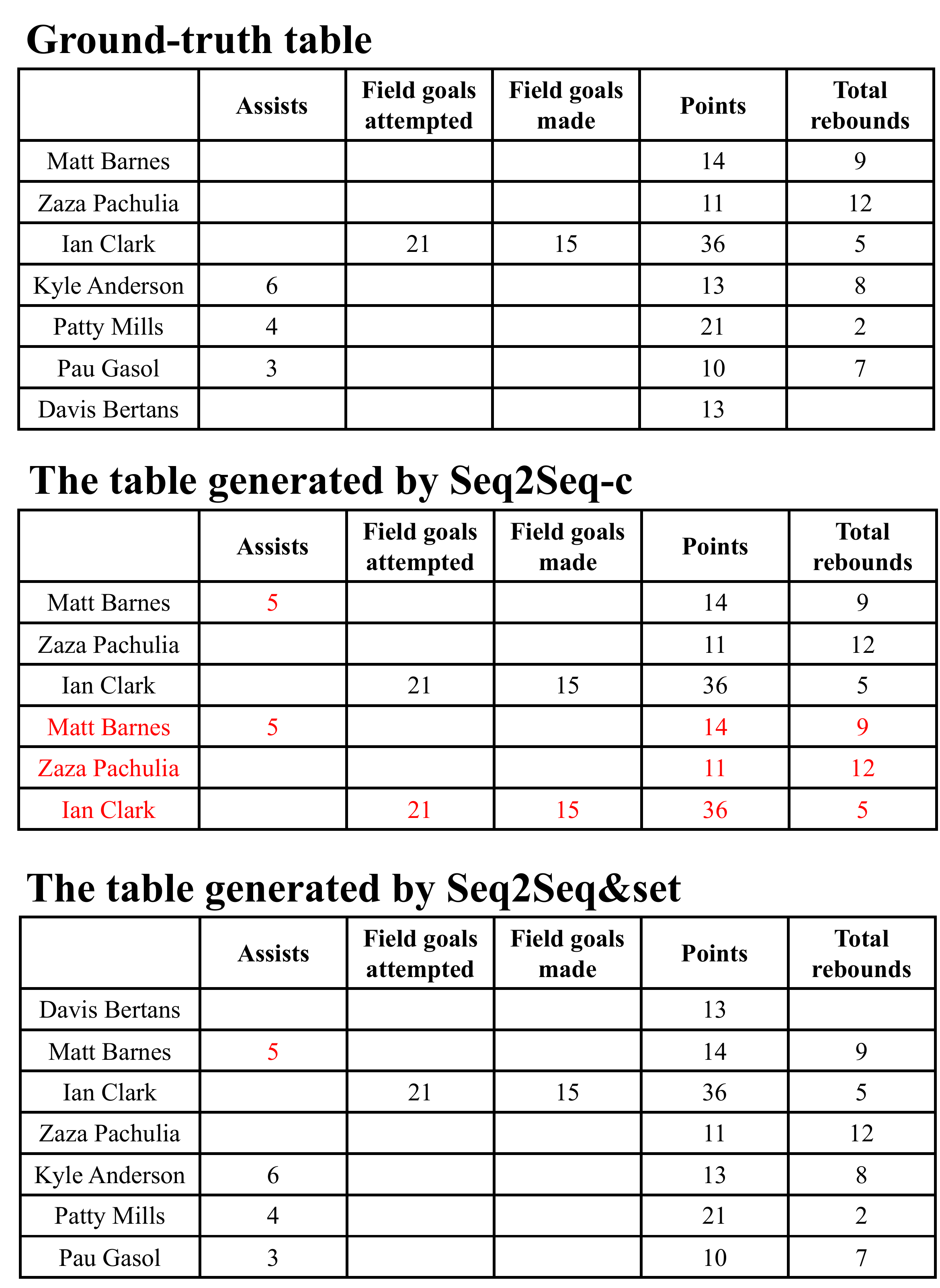}
    \caption{A case study of Seq2Seq-c and our Seq2Seq\&set model. Incorretly-generated texts are marked in red.}
    \label{fig:case} 
\end{wrapfigure}

Figure \ref{fig:case} shows a case comparison between Seq2Seq-c and our Seq2Seq\&set. Although the first three table body rows generated by Seq2Seq-c are almost correct, the others are duplicated, which also frequently occurs in other text generation tasks. In contrast, our model can handle this case correctly because ours generates table body rows in parallel and thus is not affected by other rows.

\begin{wraptable}{i}{\textwidth}
\centering
\begin{tabular}{l|ccc|c|ccc}
\toprule
Dataset & Train & Valid & Test & Avg. \# of tokens & Avg. \# of rows & Avg. \# of columns \\
\midrule 
Rotowire-Team & 3.4k & 727 & 728 & 351.05 & 2.71 & 4.84  \\
Rotowire-Player & 3.4k & 727 & 728 & 351.05 & 7.26 & 8.75  \\
E2E & 42.1k & 4.7k & 4.7k & 24.90 & 4.58 & 2.00 \\
WikiTableText & 10.0k & 1.3k & 2.0k & 19.59 & 4.26 & 2.00 \\
WikiBio & 582.7k & 72.8k & 72.7k & 122.30 & 4.20 & 2.00 \\
\bottomrule
\end{tabular}
\caption{Statistics of Rotowire, E2E, WikiTableText and WikiBio datasets, including the number of instances in training,
validation and test sets, the average number of BPE tokens per instance, and the average number of rows and columns per instance.}
\label{tab:dataset}
\end{wraptable}

\begin{wraptable}{i}{\textwidth}
\centering
\begin{tabular}{l|ccc|ccc}
\toprule
Dataset & M & $\lambda$ & $\nullrow$ scale & batch size & lr & warmup ratio \\
\midrule 
Rotowire-Team & 10 & 1 & 0.2 & 4,096 & 1e-04 & 0.01 \\
Rotowire-Player & 20 & 1 & 0.4  & 2,048 & 1e-04 & 0.01 \\
E2E & 10 & - & 0.2 & 4,096 & 1e-05 & 0.1 \\
WikiTableText & 10 & - & 0.4 & 4,096 & 1e-04 & 0.1 \\
WikiBio & 25 & - & 0.1 & 2,048 & 1e-04 & 0.1 \\
\bottomrule
\end{tabular}
\caption{The hyper-parameter settings in our experiments.}
\label{tab:hyper_parameter}
\end{wraptable}
\end{document}